\documentclass[runningheads]{llncs}

\usepackage{eccv}

\usepackage{pdflscape}
\usepackage{rotating}
\usepackage{adjustbox}

\usepackage{multirow}
\usepackage{graphicx}
\usepackage{xcolor}

\usepackage{booktabs}

\usepackage{amssymb}
\usepackage[acronym]{glossaries}
\usepackage{pifont}
\usepackage{textcomp}
\usepackage{eccvabbrv}

\usepackage{graphicx}
\usepackage{booktabs}

\usepackage[accsupp]{axessibility}  

\newcommand{\cmark}{\ding{51}}%
\newcommand{\xmark}{\ding{55}}%

\usepackage{hyperref}

\usepackage{orcidlink}
\usepackage{wrapfig}

\begin{document}



\title{Rethinking HTG Evaluation: Bridging Generation and Recognition} 

\titlerunning{Rethinking HTG Evaluation: Bridging Generation and Recognition}

\author{Konstantina Nikolaidou\inst{1}\orcidlink{0000-0002-9332-3188} \and George Retsinas\inst{2}\orcidlink{0000-0001-6734-3575} \and Giorgos Sfikas\inst{3}\orcidlink{0000-0002-7305-2886}
 \and Marcus Liwicki\inst{1}\orcidlink{0000-0003-4029-6574}}
%


\authorrunning{K.~Nikolaidou et al.}

\institute{
Luleå University of Technology, Sweden \\
\email{firstname.lastname@ltu.se}\\
\and
National Technical University of Athens, Greece\\
\email{gretsinas@central.ntua.gr}\\
\and
University of West Attica, Greece\\
\email{gsfikas@uniwa.gr}
}

\maketitle

\begin{abstract}

The evaluation of generative models for natural image tasks has been extensively studied.
Similar protocols and metrics are used in cases with unique particularities, such as Handwriting Generation, even if they might not be completely appropriate. 
In this work, we introduce three measures tailored for HTG evaluation, HTG\textsubscript{HTR}, HTG\textsubscript{style}, and HTG\textsubscript{OOV}, and argue that they are more expedient to evaluate the quality of generated handwritten images.
The metrics rely on the recognition error/accuracy of Handwriting Text Recognition and Writer Identification models and emphasize writing style, textual content, and diversity as the main aspects that adhere to the content of handwritten images.
We conduct comprehensive experiments on the IAM handwriting database, showcasing that widely used metrics such as FID fail to properly quantify the diversity and the practical utility of generated handwriting samples.
Our findings show that our metrics are richer in information and underscore the necessity of standardized evaluation protocols in HTG.
The proposed metrics provide a more robust and informative protocol for assessing HTG quality, contributing to improved performance in HTR. 
Code for the evaluation protocol is available at: \url{https://github.com/koninik/HTG_evaluation}.

\end{abstract}
\section{Introduction}
\label{sec:intro}

Handwritten Text Generation (HTG), or Styled HTG, has significantly evolved in recent years with the assistance of Deep Learning (DL) methods that have notably improved the generation quality.
The task aims to generate realistic images of readable text in a desired handwriting style, given a text prompt and a writing style condition.
HTG models not only aim to provide user personalization in digital applications but also enhance the training and performance of DL models with additional synthetic data, especially in scarce data scenarios~\cite{nikolaidou2022survey}, which are common in low-resource languages.

Existing methods introducing HTG models primarily focus on employing metrics 
designed to evaluate the quality of natural image generation~\cite{dowson1982frechet,deng2009imagenet}.
\begin{figure}
    \centering    \includegraphics[width=0.96\linewidth]{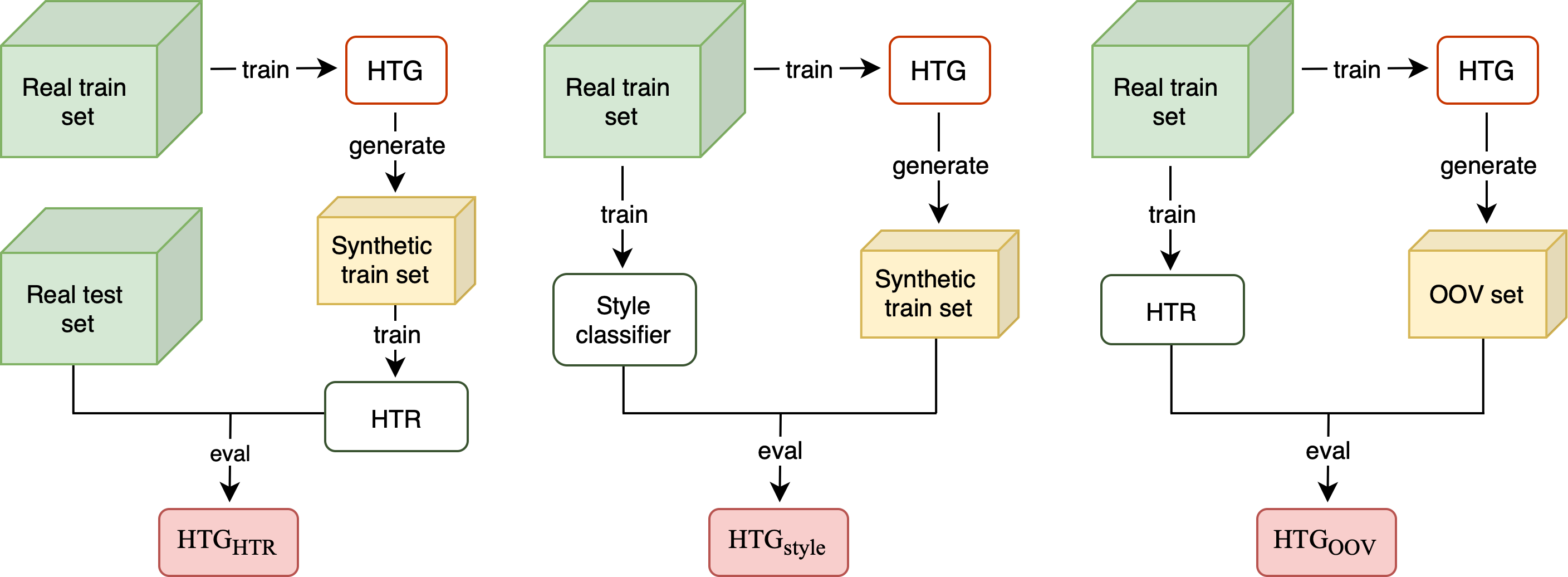}
    \caption{The proposed evaluation metrics for Handwritten Text Generation (HTG): HTG\textsubscript{HTR} (left),  HTG\textsubscript{style} (middle), and HTG\textsubscript{OOV} (right).}
    \label{fig:evaluation_protocol}
    \vspace{-2.5mm}
\end{figure}
While these metrics provide a quantitative measure in the evaluation process of generated images, 
they are not reliable or intuitive in how they measure the text or the style quality 
that is essential to define the general quality of a generated handwritten text image.
A number of works aim to address this issue by proposing novel metrics designed specifically for handwritten images~\cite{pippi2023hwd}.
However, a quantitative measure, although it plays a significant role in the evaluation of generated handwritten images, does not ensure enhanced performance and variance when integrating generated data in the training process for downstream tasks such as Handwriting Text Recognition (HTR).

When evaluating Styled HTG systems, it's crucial to consider their practical utility, particularly in enhancing the accuracy and robustness of HTR systems. 
To this end, we can identify several key properties that an effective HTG approach should possess:

\noindent
\textbf{Style Preservation.}
The ability of the HTG system to faithfully replicate the desired writing style. 
This is crucial for generating diverse and realistic handwritten text samples that can improve the robustness of HTR systems.

\noindent
\textbf{Content Preservation.}
The capability to accurately generate the requested text without introducing errors or alterations. 
This ensures the generated samples are legible and contain the correct textual information for training HTR systems.

\noindent
\textbf{Out-of-Vocabulary (OOV) Extension.}
The HTG system should be able to generate words that are not present in its training data. 
This is important for creating diverse datasets to help HTR systems generalize better to unseen words. This property is correlated to the previous two in the sense that simultaneous style and content preservation naturally lead to OOV extension. Nonetheless, this is an important property and is highlighted explicitly.

\noindent
\textbf{Variability.} 
The capacity to generate diverse samples that cover a wide range of writing styles and variations within a given style. This helps to create rich datasets that can improve the performance and generalization capabilities of HTR systems.

The majority of the existing HTG methods do not consider all these different properties, and no consistent protocol exists to take them into account and provide a spherical viewpoint of the capabilities of an HTG system.

In this paper, we propose an evaluation protocol for evaluating style- and text-conditional HTG models that go beyond generic visual feature assessment.
This protocol aims to provide a more inclusive approach to all the aspects and qualities that compose a handwritten text image by including task-driven evaluation processes covering the identified key properties.
Following this strategy, 
we rely on two respective neural network architectures that perform the tasks of Handwriting Text Recognition (HTR) and Writer Identification (WI) and propose three evaluation metrics, which we shall denote as
HTG\textsubscript{HTR}, HTG\textsubscript{style}, and HTG\textsubscript{OOV}.
Their rationale is the evaluation in terms of experimental processes that integrate synthetic-styled handwritten data as presented in~\cref{fig:evaluation_protocol}.

For HTG\textsubscript{HTR}, we train an HTR system solely on synthetic handwritten samples generated by existing HTG methods and then evaluate its performance on a real test set.
HTG\textsubscript{style} is the performance accuracy of a writing style classifier trained on real data using a subset of the real training set and testing on the rest of the corresponding generated unseen data.
The metric serves as an indication that if the trained classifier recognizes the writing style of the generated data to be as close as possible to the corresponding real data, then this is an indication that the HTG system can reproduce the writing styles of the dataset faithfully.
Furthermore, this can act as a metric of variation in the generated data, measured in terms of the various writing styles that can be generated.
Our final introduced measure, HTG\textsubscript{OOV}, focuses on the ability of the HTG system to generate OOV words, i.e., words that are not present in the training set.
The measure is computed by first training an HTR system on real training data and then generating a set of OOV words with random writing styles present in the training set using an HTG system. 
Then, the generated OOV data are evaluated in terms of the Character Error Rate (CER) performance of the HTR system. 
The idea here is that the lowest the CER is, the better the HTG system can generate correct characters in words that are not present in the real training set.




We conduct extensive experiments using the IAM database~\cite{marti2002iam} and evaluate four HTG systems~\cite{kang2020ganwriting, mattick2021smartpatch, Pippi2023HandwrittenTG,nikolaidou2023wordstylist}.
We showcase the synthetic data impact to examine the importance of variance in the generated data and the improvement of the task performance of an HTR system by incorporating filtered generated data into the process.
This action is crucial for evolving models that not only generate visually pleasing results but also generate samples that have practical applicability in downstream tasks. 
By promoting a standardized evaluation protocol, this work aims to highlight the urgent need for consistent benchmarking of HTG models that reflect the diverse applications relevant to handwriting.

The rest of the paper is structured as follows.
In~\cref{sec:related_work}, we provide a synopsis of measures and strategies used to evaluate HTG in previous works,
and present and discuss the proposed metrics in~\cref{sec:eval_protocol}.
With experimental evaluation in~\cref{sec:experiments} and related discussion in~\cref{sec:discussion}, our proposed metrics are shown to be more intuitive and informative than existing measures.

\section{HTG Evaluation}
\label{sec:related_work}


Selecting the appropriate evaluation strategy to examine the quality and practical utility of the samples generated using an HTG model is as crucial as the model training itself. 
Numerous works have presented HTG models that create handwritten text images, given a style and text condition.
To examine the quality of a generated sample, the main points that need to be considered are 
a visually pleasing result, 
the text quality (readability), 
the degree of fidelity of style imitation, 
and statistical variance in the generated data.
It is straightforward to see that these aspects cannot be trusted to be evaluated faithfully by only employing metrics designed for use in a broader, natural-image context.
These aspects can be reported through several evaluation steps using quantitative metrics specifically tailored for handwriting.
This section presents an overview of existing strategies in the literature, including standard Generative Model (GM) evaluation metrics, style evaluation, and text evaluation.
We present the aggregated information in~\cref{tab:evaluation_overview} for every included study and discuss it in the following subsections.

\subsection{Evaluation Metrics}

Several evaluation metrics are designed to assess the quality of synthetic samples generated by GMs.
Most metrics are either designed to compute similarities in the images based on specific pixel-level criteria or to compare feature similarities between the real and generated images explicitly designed for natural images~\cite{deng2009imagenet}.   
We present an overview of the evaluation metrics utilized in the HTG works presented in~\cref{tab:evaluation_overview} and comment on how appropriate each one is to evaluate samples generated using an HTG model.

\noindent
\textbf{Mean Squared Error (MSE) \& Peak Signal-to-Noise Ratio (PSNR).}
MSE measures pixel-wise differences between real and generated images, while the closely related PSNR expresses the ratio between maximum signal power and noise distortion:
\begin{align*}
\text{MSE}(\mathbf{y}, \hat{\mathbf{y}}) &= \frac{1}{n} \sum_{i=1}^{n} (y_i - \hat{y}_i)^2, \phantom{kkk}
\text{PSNR} = 10 \cdot \log_{10} \left(\frac{\text{MAX}_I^2}{\text{MSE}}\right),
\end{align*}
where $MAX_I$ denotes the maximum pixel intensity.
Due to their definition as crude pixel-wise comparison, they are not informative about the perceptual similarities in the structure and style of the images and can be prone to slight alignment differences that a human would consider insignificant.
The HTG works presented in~\cite{krishnan2023textstylebrush,zhu2023conditional} compute the MSE metric in their evaluation process but not as the primary evaluation step.
PSNR is used in the evaluation of~\cite{krishnan2023textstylebrush},~\cite{gan2022higan+}, and~\cite{nikolaidou2023wordstylist}, but is complemented with more evaluation metrics and steps.
Both metrics are designed to compute absolute errors and are thus not appropriate for capturing the degree of legibility of text or the nuanced style variations in handwritten images.

\noindent
\textbf{Geometry Score (GS).}
\textit{Geometry Score}~\cite{khrulkov2018geometry} compares the geometrical properties of the generated and learned data manifold distributions and 
is presented in several of the HTG works~\cite{alonso2019adversarial,Fogel2020ScrabbleGANSV,Davis2020TextAS,krishnan2023textstylebrush,Bhunia_2021_ICCV,luo2022slogan,Pippi2023HandwrittenTG,zhu2023conditional} along with other metrics.
GS is computed using the Mean Relative Living Times (MRLT) derived from the persistence barcodes of simplicial complexes, capturing topological features of the data. 
It is formulated as:
$$\text{GS}(X_1, X_2) = \sum_{i=0}^{i_{\max}-1} \left( \text{MRLT}(i, 1, X_1) - \text{MRLT}(i, 1, X_2) \right)^2,$$ 
where \(X_1\) and \(X_2\) are the compared datasets, and \(\text{MRLT}(i, 1, X)\) is the Mean Relative Living Times of a dataset \(X\). 
The upper bound $ i_{max}$ is related to the topological properties of the hypothesized manifolds and is, in practice, set to $100$ \cite{khrulkov2018geometry}.
While the metric can be useful for topological and structural consistencies of the generated images, it does not ensure the eligibility of the generated characters or the variations in handwriting style.

\noindent
\textbf{Structural Similarity (SSIM) Index.}
Similar to the GS score, \textit{Structural Similarity (SSIM) Index} is used to compare perceptual properties relevant to structure, as well as contrast and luminance.
The SSIM Index between two images $x$ and $y$ is computed as:
$$\text{SSIM}(x, y) = \frac{(2\mu_x \mu_y + c_1)(2\sigma_{xy} + c_2)}{(\mu_x^2 + \mu_y^2 + c_1)(\sigma_x^2 + \sigma_y^2 + c_2)},
$$ where $\mu_x$ and $\mu_y$ are the corresponding means and $\sigma_x^2$ and $\sigma_y^2$ the corresponding variances with covariance $\sigma_{xy}$ of $x$ and $y$. 
Variables $c_1$ and $c_2$ stabilize the division with weak denominators.
The works presented in~\cite{krishnan2023textstylebrush,gan2022higan+,nikolaidou2023wordstylist,zdenek2023handwritten,zhu2023conditional} utilize the metric as complementary to their evaluation strategy.
While SSIM is useful for capturing structural information and general image quality assessment, it may not fully address the intricacies when evaluating generated handwriting images. 

\noindent
\textbf{Fréchet Inception Distance (FID).}
One of the most common metrics for assessing the quality of samples from GM is the \textit{Fréchet Inception Distance (FID) score}~\cite{dowson1982frechet} defined as:
$$\text{FID} = \|\mu_{R} - \mu_{G}\|^2 + \text{Tr}(\Sigma_{R} + \Sigma_{G} - 2(\Sigma_{R}\Sigma_{G})^{1/2}).$$%
Image features of the real and generated images are extracted using an ImageNet~\cite{deng2009imagenet} pre-trained InceptionV3 network~\cite{szegedy2015goingdeeper}.
The extracted features are treated as multivariate Gaussian distribution samples with means and covariances $\mu_{R}$ and $\Sigma_{R}$ (real), and $\mu_{G}$ and $\Sigma_{G}$ (generated).
Then, the metric computes the distance between these two distributions.
While FID is used in every HTG work, as shown in~\cref{tab:evaluation_overview}, it is not suitable for handwritten images, as it is designed to evaluate natural images. 
Note also that FID acts on the statistics of the ground truth set versus the synthetic set and does not define a metric between individual elements.
FID implicitly treats the two compared sets as unimodal and Gaussian, an assumption that may often prove to be too simplistic.
Finally, FID might falsely rate a model that has memorized a dataset highly (low FID), as the two sets will have identical statistics.
An improved version of FID is the \textbf{Variable-length FID (vFID)} introduced in~\cite{Kang2021ContentAS}, where the metric operates in variable-length handwritten images.
A fine-tuned writer-style classifier on IAM database is used to extract features using a Temporal Pyramid Pooling (TPP) layer~\cite{7484698} on the convolutional features instead of the average pooling (used in FID).
Although the metric can assist in evaluating handwritten images, the metric focuses solely on the handwritten style and the code to compute vFID is not publicly available for usage.

\noindent
\textbf{Kernel Inception Distance (KID).} 
 Similar to FID, \emph{Kernel Inception Distance (KID)}~\cite{bińkowski2018demystifying} utilizes features extracted from the Inception network~\cite{szegedy2015goingdeeper} pre-trained on ImageNet~\cite{deng2009imagenet}.
 The key difference is that KID does not assume that the features follow a Gaussian distribution.
 Instead, the metric measures the distance using a mean kernel function on the Inception features and the Maximum Mean Discrepancy (MMD) approach to determine if the compared samples come from different distributions.
 It can be formally described as $\text{KID}(P, Q) = \text{MMD}^2(\phi(P), \phi(Q))$, where $\phi(P)$ and $\phi(Q)$ are the Inception representations of the samples of the examined distributions, $P$ and $Q$.
 KID is used as one of the evaluation metrics in~\cite{gan2022higan+,zdenek2023handwritten,vanherle2024vatr++}; however, it is not adequate for evaluating the variation, style, or text included in synthetic handwriting images as it is designed to evaluate natural images.

\noindent
\textbf{Inception Score (IS).}
Another popular metric to evaluate GM is the \emph{Inception Score}~\cite{salimans2016improved}.
The score deploys an Inception network~\cite{szegedy2015goingdeeper} to obtain the logits of the generated images and is formulated as: $$\text{IS} = \exp\bigg( \underset{x\sim p_G}{\mathbb{E}}\left[ D_{\text{KL}}\left(p(y|x) \,\|\, p(y)\right) \right] \bigg).
$$
Here, $p(y|x)$ denotes the conditional label distribution given an image $x$, and $p(y)$ is the marginal label distribution. 
The Kullback-Leibler divergence ($D_{\text{KL}}$) measures the difference between the two distributions. 
This metric is used in the evaluation of~\cite{gan2022higan+} along with other metrics. 
While the IS could provide insights into the diversity and quality of generated images, it is not entirely suitable for synthetic handwriting images as, importantly, 
it is in fact agnostic of the target distribution, 
and also it relies again on a network pre-trained on natural images.

\noindent
\textbf{Learned Perceptual Image Patch Similarity (LPIPS).}
LPIPS~\cite{zhang2018unreasonable} measures the perceptual similarity between image features extracted by a pre-trained deep neural network.
LPIPS is formally written as:
$$\text{LPIPS}(x, y) = \sum_{l} w_l \cdot \frac{1}{H_l W_l} \sum_{h, w} \| \phi_l(x)_{h,w} - \phi_l(y)_{h,w} \|^2,$$
where $\phi_l$ represents the feature maps of shape $H_l\times W_l$, extracted by layer $l$ of the utilized pre-trained network, normalized across the channel dimension. 
The metric has been observed to match human perception and is often used in perceptual loss networks~\cite{pihlgren2023systematic}.
Regarding HTG, LPIPS is used solely in the evaluation of~\cite{zhu2023conditional} as complementary to other metrics and experiments.
However, the used networks are again trained on natural images without including text information, thus, not making the metric suitable for evaluating handwritten images.
\noindent

\noindent
\textbf{Handwriting Distance (HWD).}
 A newly introduced work~\cite{pippi2023hwd} presents the Handwriting Distance (HWD) score specifically designed to evaluate styled handwritten images. 
 HWD computes the Euclidean distance between features extracted by a VGG16 network that is pre-trained on a large corpus of synthetic handwritten text images and is formulated as:
 \vspace{-2.5mm}
 \begin{align*}
\text{HWD} = \frac{1}{M} \sum_{m=i}^{M} \|Y_m - Y'_m\|_2, 
\end{align*}
where 
$Y_m$ and $Y'_m$
  denote the feature vectors representing real and generated images written by writer $m \in M$, respectively.
Unlike distribution-based distances, HWD captures perceptual aspects more efficiently. 
 The metric is used in~\cite{vanherle2024vatr++} to evaluate the introduced HTG system.
 While introducing this metric is an important step towards properly evaluating images generated by HTG systems, the sole use of it does not guarantee the text quality and the practical applicability of the generated data when using them in a downstream task.
\begin{table}[ht!]
\centering
\renewcommand{\arraystretch}{1.9}
\addtolength{\tabcolsep}{4pt}
\caption{Overview of existing HTG Evaluation Strategies.}
\label{tab:evaluation_overview}
\scalebox{0.55}{
\begin{tabular}{lccccc}
\toprule
\textbf{Method} & \textbf{Publication} & \textbf{GM Metrics} & \textbf{User Study} & \textbf{Style} & \textbf{HTR} \\
\bottomrule

Alonso et al.~\cite{alonso2019adversarial} & ICDAR 2019 & FID, GS & \xmark & - & \begin{tabular}[c]{@{}c@{}} HTR data augmentation
\end{tabular} \\
\hline

GANwriting~\cite{kang2020ganwriting} & ECCV 2020 & FID & \cmark & \begin{tabular}[c]{@{}c@{}}
Interpolation, t-SNE\end{tabular} & -- \\
\hline

ScrabbleGAN~\cite{Fogel2020ScrabbleGANSV} & CVPR 2020 & FID, GS &  \xmark&  
Interpolation
& \begin{tabular}[c]{@{}c@{}} HTR data augmentation\\ 
\end{tabular} \\
\hline

Davis et al.~\cite{Davis2020TextAS} & BMVC 2020 & FID, GS & \cmark & Interpolation & - \\
\hline

TextStyleBrush~\cite{krishnan2023textstylebrush} & TPAMI 2021 & \begin{tabular}[c]{@{}c@{}}FID, GS, SSIM,\vspace{-3mm} \\ PSNR, RMSE\end{tabular} & \cmark & - & - \\
\hline

Kang et al.~\cite{Kang2021ContentAS} & TPAMI 2021 & vFID & \xmark & Interpolation  &  \begin{tabular}[c]{@{}c@{}}
HTR data augmentation 
\end{tabular}\\
\hline

SmartPatch~\cite{mattick2021smartpatch} & ICDAR 2021 & FID & \cmark & - & \begin{tabular}[c]{@{}c@{}} GAN-test
\end{tabular}\\ 

\hline

HWT~\cite{Bhunia_2021_ICCV} & ICCV 2021 & FID, GS & \cmark & Interpolation& - \\
\hline

JokerGAN~\cite{zdenek2021jokergan} & ACMMM 2021 & FID & \cmark & Interpolation &  \begin{tabular}[c]{@{}c@{}}GAN-train, GAN-test\\HTR data augmentation
\end{tabular} \\
\hline

HiGAN~\cite{gan2021higan} & AAAI 2021 & FID & \xmark &  Interpolation &  \begin{tabular}[c]{@{}c@{}}HTR data augmentation
\end{tabular}\\
\hline

SLOGAN~\cite{luo2022slogan} & TNNLS 2022 & FID, GS & \cmark & \begin{tabular}[c]{@{}c@{}}Interpolation, t-SNE\end{tabular} &  \begin{tabular}[c]{@{}c@{}}HTR data augmentation
\end{tabular} \\
\hline

HiGAN+~\cite{gan2022higan+} & ACMTG 2022 & \begin{tabular}[c]{@{}c@{}}FID, IS, KID,\vspace{-3mm} \\ SSIM, PSNR\end{tabular} & \cmark &  \begin{tabular}[c]{@{}c@{}}Writer Identification Error\\Interpolation, UMAP\end{tabular}& 
\begin{tabular}[c]{@{}c@{}}
GAN-test 
\end{tabular}\\
\hline

WordStylist~\cite{nikolaidou2023wordstylist} & ICDAR 2023 & FID, SSIM, PSNR & \xmark & \begin{tabular}[c]{@{}c@{}}Writer Identification Acc\\
Interpolation, t-SNE\end{tabular} & \begin{tabular}[c]{@{}c@{}}
GAN-train \\
HTR data augmentation 
\end{tabular} \\
\hline


JokerGAN++~\cite{zdenek2023handwritten} & ICDAR 2023 & FID, KID, SSIM & \xmark & \begin{tabular}[c]{@{}c@{}}Writer Identification Error\end{tabular} & GAN-train, GAN-test \\
\hline

VATr~\cite{Pippi2023HandwrittenTG} & CVPR 2023 & FID, GS & \xmark & - & \begin{tabular}[c]{@{}c@{}}HTR data augmentation
\end{tabular} \\
\hline

CTIG-DM~\cite{zhu2023conditional} & CVPR 2023 & \begin{tabular}[c]{@{}c@{}}FID, GS, SSIM, \\ LPIPS, RMSE\end{tabular} &\xmark  & - &  \begin{tabular}[c]{@{}c@{}}HTR data augmentation\\ 
\end{tabular}\\
\hline

VATr++~\cite{vanherle2024vatr++} & arXiv 2024 & FID, KID, HWD & \xmark & - & %
\begin{tabular}[c]{@{}c@{}}
 GAN-test\\
 \end{tabular}\\

\bottomrule
\end{tabular}}
\end{table}

\subsection{Style Evaluation}

The ability of an HTG model to condition and imitate a desired writer style is a crucial aspect to evaluate when generating synthetic handwritten data.
Several works showcase this through qualitative examples of interpolating between two different writers~\cite{kang2020ganwriting,zdenek2021jokergan,gan2021higan,luo2022slogan,gan2022higan+,nikolaidou2023wordstylist}.
Additionally, a latent space visualization is often used~\cite{kang2020ganwriting,Davis2020TextAS,luo2022slogan,gan2022higan+,nikolaidou2023wordstylist,vanherle2024vatr++}.
In terms of experimental evaluation, only a few works present a style-based task to quantify style imitation~\cite{gan2022higan+,nikolaidou2023wordstylist,zdenek2023handwritten}.
HiGAN+~\cite{gan2022higan+} computes the writer identification error rate WIER using an InceptionV3 with TPP instead of average pooling and a global average pooling for the writer identifier.
The diffusion-based WordStylist~\cite{nikolaidou2023wordstylist} uses a ResNet-18~\cite{he2016deep} network fine-tuned for writer identification on the real train data and tests it on a subset of generated data to examine whether the classifier can recognize the styles.
Based on~\cite{nikolaidou2023wordstylist} and~\cite{shmelkov2018good}, we propose an appropriate evaluation protocol to evaluate HTG synthetic samples.

\subsection{Downstream HTR evaluation}

One of the primary motivations for developing HTG models is to increase the size of training samples to improve task performance.
Incorporating synthetic samples in the training process of an HTR system and subsequently evaluating the system's performance on real test images might offer multiple insights into the quality and usefulness of the generated images.
A good performance on the real test set can be interpreted as incorporating legible generated text in the training, as well as \emph{variation} in the generated styled text.

As indicated in~\cref{tab:evaluation_overview}, not all presented HTG systems include an evaluation using an HTR system.
Several works extend the training data to achieve a better HTR performance~\cite{alonso2019adversarial,Fogel2020ScrabbleGANSV,Kang2021ContentAS,zdenek2021jokergan,gan2021higan,luo2022slogan,nikolaidou2023wordstylist,Pippi2023HandwrittenTG,zhu2023conditional}.
However, it is worth noting that some works start with a relatively low baseline HTR performance, making the performance improvement more easily attainable~\cite{Fogel2020ScrabbleGANSV,Kang2021ContentAS,luo2022slogan,zhu2023conditional}. 
Moreover, other studies start with the assumption of limited data availability for the HTR training~\cite{zdenek2021jokergan}. 
In this case, if the assumption of less data is made, the generation system should also be trained with a similarly constrained dataset.
Therefore, the experimental setup plays a crucial role in the overall evaluation and outcomes of the HTG systems.

Using solely synthetic data by regenerating the training set and achieving close performance to the real corresponding data can indicate that a system can generate data that might assist the performance of an HTR system.
While this indication can be useful as a step, only three works include it in their evaluation process~\cite{zdenek2021jokergan,nikolaidou2023wordstylist,zdenek2023handwritten}.

Several works perform a GAN-test-like evaluation, where a set of the data is regenerated using the introduced HTG models~\cite{mattick2021smartpatch,zdenek2021jokergan,gan2022higan+,zdenek2023handwritten,vanherle2024vatr++}.
Then, an HTR system is trained on the real training data, and its performance is presented on the regenerated set.
While this approach, on the one hand, ensures that the generated text is readable, it is not sufficient to examine the variation of the generated data that could potentially assist an HTR system in performing better.
We further discuss the issues of this approach in~\cref{sec:eval_protocol}.

\section{Proposed Evaluation Protocol}
\label{sec:eval_protocol}

\begin{wrapfigure}{l}{7.1cm}
\vspace{-.5cm}
    \centering
\includegraphics[width=\linewidth]{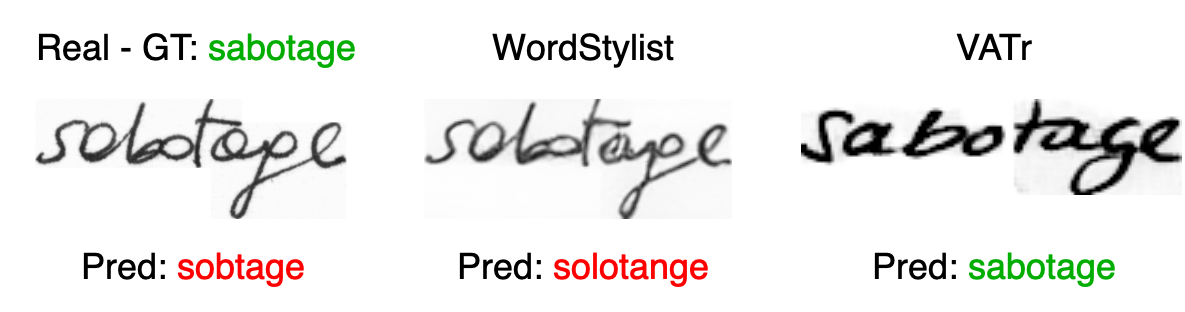}
    \caption{Issue of GAN-test evaluation.}
    \label{fig:htr_test_problem}
\vspace{-.5cm}
\end{wrapfigure}

We present an evaluation protocol for newly introduced HTG systems to promote standardization and reproducibility of the results.
The idea relies on the GAN-based evaluation presented in~\cite{shmelkov2018good}, where the GAN-train and GAN-test metrics are proposed.
GAN-train is the accuracy of a classifier trained on synthetic data and tested on real data (recall-diversity), while GAN-test is the accuracy of a classifier trained on real data and tested on synthetic data (precision-image quality).
Unlike GAN-train and GAN-test used on natural images, handwriting evaluation is more delicate as it cannot be quantified by the same task (exclusively on classification or exclusively on HTR).
As mentioned in~\cref{sec:related_work}, a few works use GAN-train~\cite{zdenek2021jokergan,nikolaidou2023wordstylist}, while others rely solely on GAN-test by regenerating the test set and evaluate it using an HTR trained on real data~\cite{mattick2021smartpatch,zdenek2021jokergan,vanherle2024vatr++}.
While this approach could showcase the ability of the model to generate understandable text that is recognizable by an HTR system, it is not useful to showcase the ability of the model to generate variations in style.
As shown in~\cite{nikolaidou2023wordstylist}, GAN-based approaches tend to generate readable text but lack style variation.
Hence, the regeneration of the test set basically simplifies the style of the text, making it easier for a well-trained HTR system to recognize and get an improved performance.
An example of this issue is presented in~\cref{fig:htr_test_problem}.
The figure depicts a real sample of the word ``sabotage'' (left) and its corresponding generated images (same style and text) from two HTG systems, WordStylist~\cite{nikolaidou2023wordstylist} (middle) and VATr~\cite{Pippi2023HandwrittenTG} (right).
Using a trained HTR system, one can see that the HTR does not correctly recognize the original real and WordStylist images due to complex style, while the VATr image contains fewer cursive connections among characters, making it easier for the HTR to correctly recognize the word.

We propose \textbf{HTG\textsubscript{HTR}}, \textbf{HTG\textsubscript{style}}, and \textbf{HTG\textsubscript{OOV}} to quantify the quality of generated images from HTG systems based on three crucial aspects for handwritten word images: the writing style, the content, and the usability of the generated images.
The proposed protocol that includes the three metrics is depicted in~\cref{fig:evaluation_protocol}.
The idea is that an HTG system is trained using a real set of images to generate different sets of synthetic images.
Three sets of experiments are conducted to compute the proposed metrics.
We present the corresponding metrics for HTG performance evaluation and their computation processes.

\noindent
\textbf{HTG\textsubscript{HTR}}.
A well-performing HTG system should be able to \emph{replicate} the variability of the training set
-- in the sense of not simply memorizing its input, 
but by creating a close variation to the input ground truth.
In other words, we want our system to provide samples characteristic of the variability 
of the true, underlying manifold as proof that it is correctly estimated.
~\cite{baek2021if} shows that less real data can be more effective than larger synthetic data sets.
Hence, evaluating how close the synthetic data behavior is to the real data can be a good indication of the quality of the generated data.
To compute HTG\textsubscript{HTR}, the synthesized training set is used to train an HTR system, which eventually will be tested on the real test set to examine how close it can reach the results that the HTR would have if it was trained on the real train set.
Hence, HTG\textsubscript{HTR} is essentially the CER percentage obtained by testing an HTR system trained on the synthetic training data on the real test set.

\noindent
\textbf{HTG\textsubscript{style}}.
Another crucial aspect to evaluate when generating style-conditioned images is 
the faithfulness of style imitation of the writers present in the dataset.
To this end, a writing style classifier is trained on a subset of the training data,
with the rest held out as a validation set.
After training, the style classifier is tested on the ``real'' evaluation set and the corresponding synthetic evaluation sets.
Similarly to the HTG\textsubscript{HTR}, 
ideally, we want to expect the HTG\textsubscript{style} metric performance
to be as close as possible to the real evaluation set.
Thus, HTG\textsubscript{style} is measured as accuracy on the generated data, similar to GAN-test.

\noindent
\textbf{HTG\textsubscript{OOV}}.
 A well-performing HTG system should also be able to generate OOV words.
 To evaluate this capability, we propose the HTG\textsubscript{OOV} metric.
 Using the introduced HTG system, we propose generating a new set that contains OOV words of random writing styles from the train set.
 Then, we evaluate an HTR system trained on the real training set on the generated OOV set and compute HTG\textsubscript{OOV} as the obtained CER (\%).
 This experiment indicates that the better/lowest the HTG\textsubscript{OOV} metric is, the more successful and clear character generation the HTG system provides when generating OOV words.

\section{Experiments}
\label{sec:experiments}

\subsection{HTG Evaluation Setup}

\noindent
\textbf{Dataset and HTG Methods.}
To compare and evaluate the HTG methods using our proposed evaluation protocol and metrics, we utilize the IAM database~\cite{marti2002iam}, the most commonly used dataset for HTR and HTG.
Among the wide variety of HTG systems that exist in the literature, as can also be seen from~\cref{tab:evaluation_overview}, we present our evaluation using a few representatives of them to cover several architectures, such as GANs~\cite{kang2020ganwriting,mattick2021smartpatch}, Transformers~\cite{Pippi2023HandwrittenTG}, and Diffusion models~\cite{nikolaidou2023wordstylist}.

\noindent
\textbf{Metrics Implementation.}
For our proposed HTG\textsubscript{HTR}, we use the state-of-the-art HTR system presented in~\cite{retsinas2022best} that uses a standard CNN-LSTM architecture but proposes best practices to maximize the CRNN performance.
For the HTG\textsubscript{style}, we utilize an off-the-shelf ResNet-18 pretrained on ImageNet and train it on 70\% of the training set, keeping the rest 30\%  as a validation set to compute the metric.
To create this validation set, we use a random split.
We provide the corresponding split for reproducibility.
Compared to our proposed metrics, we compute two of the most frequently used evaluation metrics, FID and KID, which are based on features extracted by ImageNet pre-trained networks.
We include the newly introduced HWD~\cite{pippi2023hwd} designed for handwritten images.
All three metrics are computed using the code provided by the HWD~\cite{pippi2023hwd} work.

\subsection{Evaluation Results}

\begin{table}[t]
\centering
  \renewcommand{\arraystretch}{1.3}
  \addtolength{\tabcolsep}{3.5pt}
  \caption{Evaluation of HTG methods using our proposed metrics HTG\textsubscript{HTR}, HTG\textsubscript{style}, and HTG\textsubscript{OOV}. We further compare include FID, KID, and HWD.}
  \scalebox{0.75}{
\begin{tabular}{lcccccc}
\toprule
\textbf{Method}      & \textbf{FID\textdownarrow} & \textbf{KID\textdownarrow} & \textbf{HWD\textdownarrow} & \textbf{HTG\textsubscript{HTR}\textdownarrow} & \textbf{HTG\textsubscript{style}\textuparrow} &
\textbf{HTG\textsubscript{OOV}\textdownarrow} \\
\bottomrule

real images & -   & -  & -   & 5.14     & 82.05 & - \\
\midrule
GANwriting~\cite{kang2020ganwriting}  & 37.41   & 0.0196  & 0.610   & 39.56       &   4.59   & 7.45 \\
SmartPatch~\cite{mattick2021smartpatch} & 48.24   & 0.0331  & 0.641   & 39.22       &   3.00    & 9.20 \\
VATr~\cite{Pippi2023HandwrittenTG}& 27.79   & 0.0105  & 0.591  & 21.37       &     1.39   & 5.42\\
WordStylist~\cite{nikolaidou2023wordstylist} & 36.69   & 0.0194  & 0.303   & 8.23   & 67.12  & 29.85  \\      
\bottomrule
\end{tabular}}
\label{tab:metric_results}
\end{table}
We provide insight into the proposed metrics by examining the results presented in~\cref{tab:metric_results}.
HTG\textsubscript{HTR} results reveal its relation to the variance of the generated data.
As can also be seen in the evaluation of~\cite{nikolaidou2023wordstylist}, WordStylist has the lowest HTG\textsubscript{HTR} score of 8.23\%, showcasing the highest variance among the generated data and the closest one to the real data variance which is a crucial aspect when training an HTR system.
VATr~\cite{Pippi2023HandwrittenTG} shows the next best HTG\textsubscript{HTR} of 21.37\%, with GANwriting and SmartPatch following with a high HTG\textsubscript{HTR} of $\sim$ 39\%.
HTG\textsubscript{HTR} seems to align with the HWD metric while showing similar behavior in terms of FID and KID, but with VATr having the best results in these two metrics.
It should be noted that for VATr, the FID, KID, and HWD metrics are computed similarly to the original paper for fairness to the method, where the computations occur between the generated data and a processed version of the real data that the model saves as the generation occurs.

HTG\textsubscript{style} reveals a similar trend to HTG\textsubscript{HTR}, with WordStylist showing the best performance in imitating the writing style of the dataset, while the rest of the methods are not as successful.
WordStylist achieves a score of 67.12\% HTG\textsubscript{style}, a 15\% lower score than the real data performance of 82.05\%.
GANwriting, SmartPatch, and VATr achieve less than 5\%, showing a weak style imitation in their generated data.

HTG\textsubscript{OOV} shows entirely different behavior on the examined HTG methods.
WordStylist has a high value of 29.85\%, while the other methods achieve less than 10\%, with VATr showcasing the best result of 5.42\%.
This indicates that VATr generates the most accurate word images in terms of text condition, while WordStylist seems to create a high amount of noisy text that is not perfectly recognizable.
Despite the noise, there is still a useful amount of generated data, as shown in~\cref{sub:practical_utility} and~\cref{fig:barplot}.  

\subsection{Synthetic Data Impact on HTG\textsubscript{HTR}}

\begin{figure}
    \centering    \includegraphics[width=0.81\linewidth]{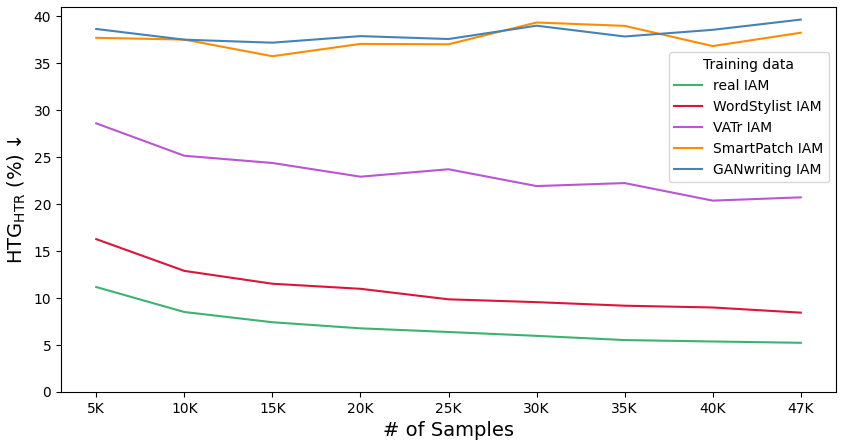}
    \caption{Impact of gradually adding synthetic data
of the examined HTG methods to the training process of HTG\textsubscript{HTR} metric until the original IAM training set size is reached.}
    \label{fig:HTG_HTR}
\end{figure}

We examine the impact of the synthetic data by gradually adding more synthetic samples to the training process of HTG\textsubscript{HTR}.
The intuition behind this is that the more data is added to the training process, the more variation is ``injected''.
A good HTG system should generate samples that asymptotically cover the true manifold of handwritten words (of which the training set represents a sample) and not just repeat its input with limited (or no) variability.
Hence, we expect to observe a gradual improvement in the HTG\textsubscript{HTR} metric by adding more synthetic data. 
For every HTG method, we start by training the baseline HTR system~\cite{retsinas2022best} that computes the HTG\textsubscript{HTR} metric with 5K synthetic samples and no real data and repeat the experiment by adding an extra 5K samples in every step of the process until we reach the size of the real training set which is $\sim$ 47K.
We present the experimental results of the HTG systems and compare them to the behavior of the real training data in~\cref{fig:HTG_HTR}.

~\cref{fig:HTG_HTR} shows an expected behavior if we consider the HTG\textsubscript{HTR} results of~\cref{tab:metric_results}.
WordStylist shows the closest behavior to the real training data; however, there is still room for improvement to reach the quality of the real data.
This signifies that the method is not able to provide as much variation as the real data, or, while considering the result of the HTG\textsubscript{OOV} metric in~\cref{tab:metric_results}, the system is prone to generate noisy data or incorrect characters.
VATr shows a similar drop in the HTG\textsubscript{HTR} as more data is added to the training process.
However, the plot shows some small instabilities while the values of HTG\textsubscript{HTR} are quite higher than those of WordStylist or the real training data.
Finally, GANwriting and SmartPatch show no improvement when additional data is used and instabilities that reveal the lack of variation in the generated data.

\subsection{Practical Utility of the Generated Data}
\label{sub:practical_utility}
We examine the practical utility of the generated data and show whether their addition improves the performance of an HTR system.
We examine two HTG system cases.
One case includes GANwriting~\cite{kang2020ganwriting} that shows a low data variation and style imitation according to HTG\textsubscript{HTR} and HTG\textsubscript{style} and a successful character generation according to HTG\textsubscript{OOV}.
The other case includes WordStylist~\cite{nikolaidou2023wordstylist}, a system showing high data variance and style replication and a high value of HTG\textsubscript{OOV}.
Using both HTG methods, we generate a large corpus of 180K samples that include both IV and OOV words and a random writing style of the IAM training set.
The generated corpus from each system is used as additional training data to the real data to improve the performance of the HTR system and examine the usability of the generated data.
Similarly to previous experiments, we use~\cite{retsinas2022best} as our baseline HTR system.

\begin{figure}
    \centering    
    \includegraphics[width=0.8\linewidth]{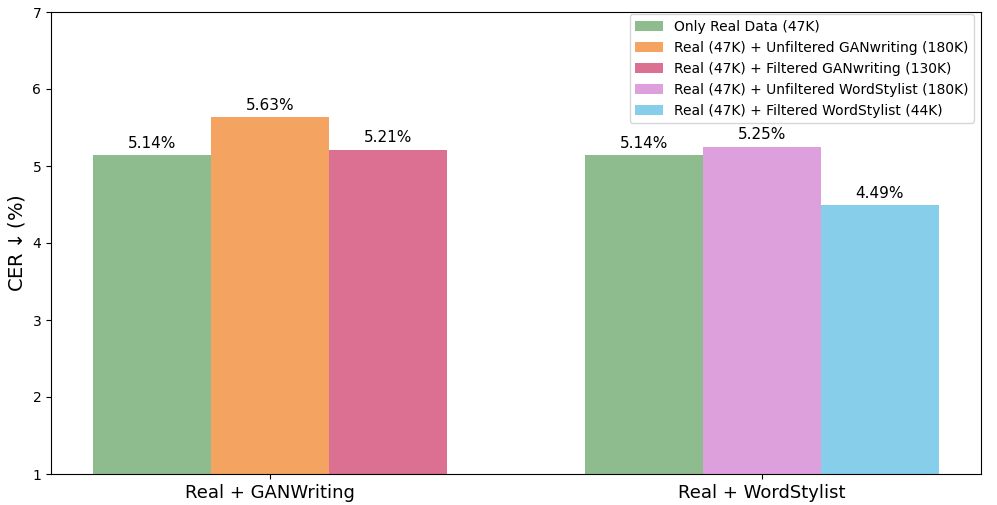}
    \caption{Impact of filtered and unfiltered synthetic data on HTR performance on the real test set of IAM database. The left group presents data generated by GANwriting, and the right group is generated by WordStylist. Both groups are compared with the performance when training only using the real training set.}
    \label{fig:barplot}
\end{figure}
Considering the HTG\textsubscript{OOV} results, we further examine how filtering the data can assist their utility in the downstream task.
The filtering occurs by testing the generated corpus using the HTR system trained only on the real data and keeping only the generated data with a CER of 0, meaning all characters are correctly generated.
The filtering process keeps 130K ``clean'' samples for GANwriting and only 44K for WordStylist, which was expected according to their HTG\textsubscript{OOV} values.

We present the experiments on the filtered and unfiltered data in~\cref{fig:barplot}, where we include the following five scenarios of training the HTR system: only with real data, with the combination of real data and unfiltered or filtered GANwriting-generated data, and the combination of real data and unfiltered or filtered WordStylist-generated data.
As we can observe from the results, both unfiltered data cases show worse performance than the real data alone, although the amount of training data has increased.
When filtering the large corpus, we can see that, in the case of GANwriting, still GANwriting data cannot assist the performance of the HTR system. 
However, the performance is better than that of the filtered GANwriting data.  
In the case of WordStylist, the filtered data improved the HTR performance with a CER of 4.49\%, although the filtering process has kept roughly 25\% of the generated data.

\section{Discussion and Conclusion}
\label{sec:discussion}

We presented an evaluation protocol using HTG\textsubscript{HTR}, HTG\textsubscript{style}, and HTG\textsubscript{OOV} metrics to assess the quality of HTG systems by evaluating the variance, the style imitation, and the successful content generation of synthetic data.
The protocol can serve as a guideline for future research to establish a standardized process to evaluate synthetic handwritten text samples depending on the dataset needs.
We conducted experiments focusing on the variability and practical utility of generated data to support the proposed evaluation metrics.
Here, we highlight key observations obtained while introducing the proposed metrics and showcasing the experimental results.

\noindent
\textbf{Limitations of GM Metrics.}
Although the same GM metrics are used in most works, they are insufficient to examine the quality of the generated handwritten data as they either focus on natural image features or pixel-level structural properties.
For example, GANwriting and WordStylist have very close values in terms of FID and KID, as shown in~\cref{tab:metric_results}, while our proposed metrics and practical utility experiments show a big difference in the two HTG systems.   

\noindent
\textbf{Lack of Standardized Protocol.}
While some works focus on experimental evaluation of the generated data considering the key properties of handwritten generation, there is no standardized protocol to properly follow the processes.

\noindent
\textbf{Importance of Data Variability.}
Our experimental results showcase the importance of data variation in improving downstream task performance, where our proposed HTG\textsubscript{HTR} and HTG\textsubscript{style} seem to give intuition.

\noindent
\textbf{Practical Utility of Generated Data.}
The results of~\cref{sub:practical_utility} validate the intuition and usefulness of our proposed metrics.
Bad values of HTG\textsubscript{HTR} (high) and HTG\textsubscript{style} (low) signify a low style variation in the generated data.
Similarly, bad values of HTG\textsubscript{OOV} (high) show the inability of the HTG to extend to a larger dataset.
This can be mitigated by filtering readable data through an HTR system and keeping ``clean'' data that can assist the training process.
However, this process is computationally expensive as it requires the generation of more data to keep a sufficient amount of successfully generated samples.
    
\noindent
\textbf{Limitations and Future Work.}
Given our results, there is plenty of room for improvement in developing HTG systems that cover all key properties. 
Our proposed metrics, while informative on the handwriting properties, rely on external models, which increases computation.
Many works also evaluate unseen generated styles, not addressed here, as we focus on bridging generation with handwriting recognition. 
Fine-tuning the HTG\textsubscript{style} classifier could effectively evaluate these unseen styles.
Future possibilities could focus on refining the proposed metrics with an more tailored network design for handwriting, or reducing reliance on external models.
Finally, expanding the protocol to more experiments and including diverse datasets will enhance its generalizability.

To conclude, our work highlights the urgent need for proper evaluation of data created by HTG systems and aims to standardize an evaluation protocol for HTG.
By providing metrics richer in information, we aim to promote the introduction of HTG systems that are not only visually pleasing but can also contribute to the improvement of HTR performance.

\section*{Acknowledgment}
\noindent
The computations and data handling were enabled by the Berzelius resource provided by the Knut and Alice Wallenberg Foundation at the National Supercomputer Centre at Linköping University.

%
%
\bibliographystyle{splncs04}
\bibliography{main}

\begin{thebibliography}{10}
\providecommand{\url}[1]{\texttt{#1}}
\providecommand{\urlprefix}{URL }
\providecommand{\doi}[1]{https://doi.org/#1}

\bibitem{alonso2019adversarial}
Alonso, E., Moysset, B., Messina, R.: {Adversarial Generation of Handwritten Text Images Conditioned on Sequences}. In: 2019 International Conference on Document Analysis and Recognition (ICDAR). pp. 481--486. IEEE (2019)

\bibitem{baek2021if}
Baek, J., Matsui, Y., Aizawa, K.: What if we only use real datasets for scene text recognition? toward scene text recognition with fewer labels. In: Proceedings of the IEEE/CVF conference on computer vision and pattern recognition. pp. 3113--3122 (2021)

\bibitem{Bhunia_2021_ICCV}
Bhunia, A.K., Khan, S., Cholakkal, H., Anwer, R.M., Khan, F.S., Shah, M.: {Handwriting Transformers}. In: Proceedings of the IEEE/CVF International Conference on Computer Vision (ICCV). pp. 1086--1094 (October 2021)

\bibitem{bińkowski2018demystifying}
Bińkowski, M., Sutherland, D.J., Arbel, M., Gretton, A.: Demystifying {MMD} {GAN}s. In: International Conference on Learning Representations (2018)

\bibitem{Davis2020TextAS}
Davis, B.L., Tensmeyer, C., Price, B.L., Wigington, C., Morse, B., Jain, R.: {Text and Style Conditioned GAN for the Generation of Offline-Handwriting Lines}. ArXiv  \textbf{abs/2009.00678} (2020)

\bibitem{deng2009imagenet}
Deng, J., Dong, W., Socher, R., Li, L.J., Li, K., Fei-Fei, L.: {ImageNet: A Large-Scale Hierarchical Image Database}. In: 2009 IEEE conference on computer vision and pattern recognition. pp. 248--255. Ieee (2009)

\bibitem{dowson1982frechet}
Dowson, D., Landau, B.: {The Fr{\'e}chet distance between multivariate normal distributions}. Journal of Multivariate Analysis  \textbf{12}(3),  450--455 (1982)

\bibitem{gan2021higan}
Gan, J., Wang, W.: {HiGAN: Handwriting Imitation Conditioned on Arbitrary-Length Texts and Disentangled Styles}. In: Proceedings of the AAAI Conference on Artificial Intelligence. vol.~35, pp. 7484--7492 (2021)

\bibitem{gan2022higan+}
Gan, J., Wang, W., Leng, J., Gao, X.: {HiGAN+: Handwriting Imitation GAN with Disentangled Representations}. ACM Transactions on Graphics (TOG)  \textbf{42}(1),  1--17 (2022)

\bibitem{he2016deep}
He, K., Zhang, X., Ren, S., Sun, J.: {Deep Residual Learning for Image Recognition}. In: Proceedings of the IEEE Conference on Computer Vision and Pattern recognition. pp. 770--778 (2016)

\bibitem{Kang2021ContentAS}
Kang, L., Riba, P., Rusi{\~n}ol, M., Forn{\'e}s, A., Villegas, M.: {Content and Style Aware Generation of Text-Line Images for Handwriting Recognition}. IEEE Transactions on Pattern Analysis and Machine Intelligence  \textbf{44},  8846--8860 (2021)

\bibitem{kang2020ganwriting}
Kang, L., Riba, P., Wang, Y., Rusinol, M., Forn{\'e}s, A., Villegas, M.: {GANwriting: Content-Conditioned Generation of Styled Handwritten Word Images}. In: European Conference on Computer Vision. pp. 273--289. Springer (2020)

\bibitem{khrulkov2018geometry}
Khrulkov, V., Oseledets, I.: {Geometry Score: A Method for Comparing Generative Adversarial Networks}. In: International Conference on Machine Learning. pp. 2621--2629. PMLR (2018)

\bibitem{krishnan2023textstylebrush}
Krishnan, P., Kovvuri, R., Pang, G., Vassilev, B., Hassner, T.: {TextStyleBrush: Transfer of Text Aesthetics from a Single Example}. IEEE Transactions on Pattern Analysis and Machine Intelligence  (2023)

\bibitem{luo2022slogan}
Luo, C., Zhu, Y., Jin, L., Li, Z., Peng, D.: {SLOGAN: Handwriting Style Synthesis for Arbitrary-Length and Out-of-Vocabulary Text}. IEEE Transactions on Neural Networks and Learning Systems  (2022)

\bibitem{marti2002iam}
Marti, U.V., Bunke, H.: {The IAM-database: an English sentence database for offline handwriting recognition}. International Journal on Document Analysis and Recognition  \textbf{5},  39--46 (2002)

\bibitem{mattick2021smartpatch}
Mattick, A., Mayr, M., Seuret, M., Maier, A., Christlein, V.: Smartpatch: Improving handwritten word imitation with patch discriminators. In: International Conference on Document Analysis and Recognition. pp. 268--283. Springer (2021)

\bibitem{nikolaidou2023wordstylist}
Nikolaidou, K., Retsinas, G., Christlein, V., Seuret, M., Sfikas, G., Smith, E.B., Mokayed, H., Liwicki, M.: {WordStylist: Styled Verbatim Handwritten Text Generation with Latent Diffusion Models}. In: International Conference on Document Analysis and Recognition. pp. 384--401. Springer (2023)

\bibitem{nikolaidou2022survey}
Nikolaidou, K., Seuret, M., Mokayed, H., Liwicki, M.: A survey of historical document image datasets. International Journal on Document Analysis and Recognition (IJDAR)  \textbf{25}(4),  305--338 (2022)

\bibitem{pihlgren2023systematic}
Pihlgren, G.G., Nikolaidou, K., Chhipa, P.C., Abid, N., Saini, R., Sandin, F., Liwicki, M.: {A Systematic Performance Analysis of Deep Perceptual Loss Networks: Breaking Transfer Learning Conventions}. arXiv preprint arXiv:2302.04032  (2023)

\bibitem{Pippi2023HandwrittenTG}
Pippi, V., Cascianelli, S., Cucchiara, R.: {Handwritten Text Generation from Visual Archetypes}. In: Proceedings of the IEEE/CVF Conference on Computer Vision and Pattern Recognition. pp. 22458--22467 (2023)

\bibitem{pippi2023hwd}
Pippi, V., Quattrini, F., Cascianelli, S., Cucchiara, R.: {HWD: A Novel Evaluation Score for Styled Handwritten Text Generation}. arXiv preprint arXiv:2310.20316  (2023)

\bibitem{retsinas2022best}
Retsinas, G., Sfikas, G., Gatos, B., Nikou, C.: Best practices for a handwritten text recognition system. In: International Workshop on Document Analysis Systems. pp. 247--259. Springer (2022)

\bibitem{salimans2016improved}
Salimans, T., Goodfellow, I., Zaremba, W., Cheung, V., Radford, A., Chen, X.: {Improved Techniques for Training GANs}. Advances in Neural Information Processing Systems  \textbf{29} (2016)

\bibitem{Fogel2020ScrabbleGANSV}
{Sharon Fogel and Hadar Averbuch-Elor and Sarel Cohen and Shai Mazor and Roee Litman}: {ScrabbleGAN: Semi-Supervised Varying Length Handwritten Text Generation}. {2020 IEEE/CVF Conference on Computer Vision and Pattern Recognition (CVPR)} pp. {4323--4332} ({2020})

\bibitem{shmelkov2018good}
Shmelkov, K., Schmid, C., Alahari, K.: {How good is my GAN?} In: Proceedings of the European Conference on Computer Vision (ECCV). pp. 213--229 (2018)

\bibitem{szegedy2015goingdeeper}
Szegedy, C., Liu, W., Jia, Y., Sermanet, P., Reed, S., Anguelov, D., Erhan, D., Vanhoucke, V., Rabinovich, A.: {Going Deeper with Convolutions}. In: Proceedings of the IEEE Conference on Computer Vision and Pattern Recognition. pp.~1--9 (2015)

\bibitem{vanherle2024vatr++}
Vanherle, B., Pippi, V., Cascianelli, S., Michiels, N., Van~Reeth, F., Cucchiara, R.: {VATr++: Choose Your Words Wisely for Handwritten Text Generation}. arXiv preprint arXiv:2402.10798  (2024)

\bibitem{7484698}
Wang, P., Cao, Y., Shen, C., Liu, L., Shen, H.T.: {Temporal Pyramid Pooling-Based Convolutional Neural Network for Action Recognition}. IEEE Transactions on Circuits and Systems for Video Technology  \textbf{27}(12),  2613--2622 (2016)

\bibitem{zdenek2021jokergan}
Zdenek, J., Nakayama, H.: {JokerGAN: memory-efficient model for handwritten text generation with text line awareness}. In: Proceedings of the 29th ACM international conference on multimedia. pp. 5655--5663 (2021)

\bibitem{zdenek2023handwritten}
Zdenek, J., Nakayama, H.: Handwritten text generation with character-specific encoding for style imitation. In: International Conference on Document Analysis and Recognition. pp. 313--329. Springer (2023)

\bibitem{zhang2018unreasonable}
Zhang, R., Isola, P., Efros, A.A., Shechtman, E., Wang, O.: {The Unreasonable Effectiveness of Deep Features as a Perceptual Metric}. In: Proceedings of the IEEE Conference on Computer Vision and Pattern Recognition. pp. 586--595 (2018)

\bibitem{zhu2023conditional}
Zhu, Y., Li, Z., Wang, T., He, M., Yao, C.: {Conditional Text Image Generation with Diffusion Models}. In: Proceedings of the IEEE/CVF Conference on Computer Vision and Pattern Recognition. pp. 14235--14245 (2023)

\end{thebibliography}
\end{document}